%% file: acl2023.tex
\pdfoutput=1

\documentclass[11pt]{article}

\usepackage{ACL2023}

\usepackage{tcolorbox}
\tcbset{
  boxrule=0.4pt,
  arc=1.5mm,
  left=2mm,
  right=2mm,
  top=1mm,
  bottom=1mm,
  colframe=gray!60!black
}

\usepackage{times}
\usepackage{latexsym}
\usepackage{graphicx}
\usepackage{xcolor}
\usepackage{fancyvrb}
\usepackage{framed}

\usepackage[T1]{fontenc}

\usepackage[utf8]{inputenc}

\usepackage{microtype}

\usepackage{inconsolata}

\usepackage{multirow}
\usepackage{graphicx}
\usepackage{moreverb}
\usepackage{amsmath}
\usepackage{amsfonts}
\usepackage{tablefootnote} 
\usepackage{todonotes}

\usepackage{cleveref}

\title{
Event Detection with a Context-Aware Encoder and LoRA for Improved Performance on Long-Tailed Classes

}

\author{
  Abdullah Al Monsur \\
  University of South Florida \\
  \texttt{almonsur@usf.edu}
  \And
  Nitesh Vamshi Bommisetty \\
  University of South Florida \\
  \texttt{bn319@usf.edu}
  \And
  Gene Louis Kim \\
  University of South Florida \\
  \texttt{genekim@usf.edu}
}

\begin{document}
\maketitle
\begin{abstract}
The current state of event detection research has two notable re-occurring limitations that we investigate in this study. First, the unidirectional nature of decoder-only LLMs presents a fundamental architectural bottleneck for natural language understanding tasks that depend on rich, bidirectional context. Second, we confront the conventional reliance on Micro-F1 scores in event detection literature, which systematically inflates performance by favoring majority classes. Instead, we focus on Macro-F1 as a more representative measure of a model's ability across the long-tail of event types. Our experiments demonstrate that models enhanced with sentence context achieve superior performance over canonical decoder-only baselines. Using Low-Rank Adaptation (LoRA) during finetuning provides a substantial boost in Macro-F1 scores in particular, especially for the decoder-only models, showing that LoRA can be an effective tool to enhance LLMs' performance on long-tailed event classes.
\end{abstract}

%
%

\section{Introduction}
\label{sec:introduction}

\input{sections/introduction}

\section{Related Work}
\label{sec:related_work}
\input{sections/related_work}

\section{Methodology}
\label{sec:methodology}
\input{sections/methodology}

\section{Experimental Setup}
\label{sec:experimental_setup}

\input{sections/experimental_setup}

\section{Results}

\label{sec:results}
\input{sections/results}

\section{Conclusion}
\label{sec:conclusion}
\input{sections/conclusion}

\section*{Limitations}


\textbf{Error and difficulty analysis.} A breakdown by context length and type rarity, showing where improvements occur, would give us a better idea of the function of the models.

\textbf{Prompt technique exploration.} We did not investigate advanced prompt-based or instruction-based methods in depth, so their impact on macro‑averaged metrics was not explored. 

\textbf{Hyperparameter tuning.} Hyperparameter search was necessarily bounded; a wider sweep exploration might yield additional gains.  

\textbf{Testing on multiple splits.} We used a single data split. Testing multiple random splits and seeds would provide a clearer assessment of variance and outcome consistency.

\textbf{Comparing the results against the latest and best models.} The results were not compared with the other latest and state-of-the-art models. As approaches and evaluation techniques differ for each model and framework, comparing the results of our model with the other ones requires reproducing the results in a unified framework. It was not done in this project.

\bibliography{anthology_1, custom}
\bibliographystyle{acl_natbib}

%

\newpage

\appendix

\input{appendices/default_hyperparameters}

\end{document}

%% file: sections/introduction.tex
The task of event detection~(ED) is to identify and categorize the events in natural language~\cite{grishman-1997-information,chinchor-marsh-1998-appendix,ahn-2006-stages}. This forms an important basis for structured interpretation of text and downstream applications that focus on event information~\cite{zhang-etal-2020-aser,han-etal-2021-ester}.

The event detection literature is dominated by the encoder-only models, which are 
typically based on the outdated BERT models~\cite{huang-etal-2024-textee}. This stands in contrast to the decoder-only models that dominate the current LLM landscape, such as Llama~\cite{grattafiori2024llama3herdmodels} and Qwen~\cite{bai2023qwentechnicalreport}. 
The pretraining on vast corpora has led to remarkable generative capabilities, but their unidirectional attention mechanism has been a roadblock in getting strong performance on embedding-based uses that access token embeddings before full access to the context processing~\cite{saattrup-nielsen-etal-2025-encoder, dukic-snajder-2024-looking}. We aim to address this issue in the context of event detection. We hypothesize that whole sentence context is critical for generalizable event detection performance, enabling models to distinguish underrepresented events more accurately. These event types with few training examples require differentiation of subtle variations with limited information and thus the full sentence context is likely needed.

\begin{figure}[t!]
    \centering
    \includegraphics[width=\linewidth]{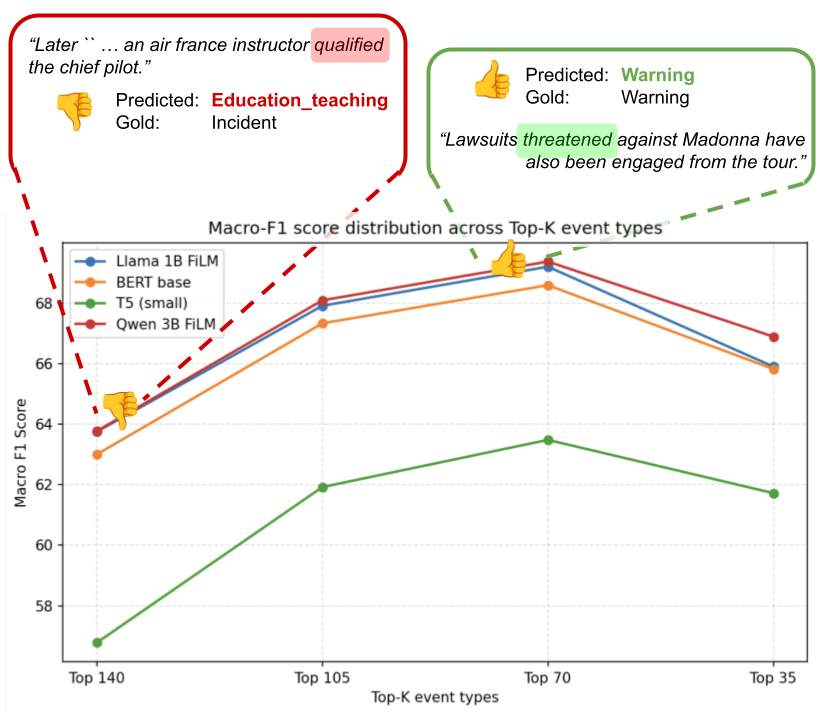}
    \caption{Macro-F1 scores across quartiles of event types, ordered by event mention frequency for four models, BERT, T5, Llama 1B FiLM, and Qwen 3B FiLM. We find that models consistently underperform on the events with fewest mentions (left side of the plot). The top of the diagram shows examples of high-frequency event, \texttt{Warning}, with a correct prediction and low-frequency event, \texttt{Incident}, with an incorrect prediction from Qwen 3B FiLM.}
    \label{fig:intro-figure}
\end{figure}


Looking at evaluation, most recent event detection research use Micro-F1 scores, which hides generalizability of the model to long-tailed classes~\cite{huang-etal-2024-textee}. 
In this paper, we make the case that Macro-F1 is an important complementary metric to assess the competence of models on long-tailed classes. The importance of Macro-F1 is highlighted by the divergence of model rankings in our paper when using the two metrics. 

Recent advances in parameter-efficient fine-tuning, such as LoRA, have shown promise in mitigating forgetting during adaptation. Our intuition is that this might help the LLMs to generalize more across the classes, improving the long-tailed class performance. Also, large models might overfit to the training data due to huge number of parameters. As a result, they might not perform well on the test or validation set. LoRA might be able to mitigate this issue. 

In-context learning is at times fast and efficient because it does not require training the model. Some might prefer it over the supervised finetuning (SFT) approaches due to this factor. To evaluate how SFT differs in performance from in-context learning, we conducted few-shot and zero-shot experiments.

Our research questions are:

\textbf{RQ1:} What benefits, if any, are gained by introducing sentence-level information to decoder-only models for event detection? And which methods of doing so provide the greatest benefits?

\textbf{RQ2:} Does Macro-F1 capture a distinct and important aspect of event detection performance that is overlooked when only using Micro-F1 scores?

\textbf{RQ3:} How does LoRA impact event detection performance along both Micro- and Macro-F1 metrics?


Our key contributions of this paper are the following:
\begin{itemize} 
    \item We demonstrate that the performance of decoder-only LLMs for event detection can be significantly improved by incorporating sentence-level information during the finetuning process. The feature-wise linear modulation~(FiLM; \Cref{sssec:film}) method was overall most effective among them.
    

    \item Event detection methods show distinct Micro- and Macro-F1 performance behaviors, motivating the need to include Macro-F1 for complete comparisons.
    
    \item We find that beyond computational efficiency, LoRA acts as a potent regularizer that enhances generalization to low-frequency event types, leading to substantial gains in Macro-F1 scores across nearly all experimental setups.
    
\end{itemize}

%% file: sections/related_work.tex
The introduction of the transformer architecture and large-scale pretrained language models 
altered the landscape of event detection. Models like BERT not only established a new state-of-the-art in performance but also inspired a conceptual shift in how the task itself was approached, moving from simple classification to more sophisticated, semantically-grounded paradigms~\cite{hettiarachchi-ranasinghe-2023-explainable}. Latest powerful decoder-only LLMs, have not yet been deeply investigated over their BERT-based counterparts on event detection benchmarks~\cite{huang-etal-2024-textee}.

The MAVEN~\cite{wang-etal-2020-maven} dataset represents a significant leap forward in scale of event detection datasets, constructed from English Wikipedia articles. The RAMS~\cite{ebner-etal-2020-multi} is made from news articles which is another dataset with large number of event types.


LoRA offers a more computationally efficient method of finetuning LLMs by reducing the number of trainable parameters via low-rank matrices. This method allows the modification of every parameter of the original network without the full computational cost and has shown to achieve competitive performance to full-parameter finetuning~\cite{hu-etal-2022-lora}.


Prior research has aimed to mitigate the shortcomings of decoder-only models. LLM2Vec~\cite{behnamghader2024llm2veclargelanguagemodels} and Dec2Enc~\cite{HUANG2025112907} are the two existing  methods that introduce bidirectional attention to the decoder-only models. These methods have been evaluated across a wide range of tasks, but event detection evaluations are notably lacking.

The evaluation of ED and its sub-tasks has long been standardized around the metrics of Precision, Recall, and F1-score, which are micro-averaged ~\cite{simon-etal-2024-generative}. This poses a limitation of overlooking the model performance on long-tailed classes which are underrepresented in the dataset, and triggers the need to evaluate using macro-averaged metrics.

%% file: sections/methodology.tex
\begin{figure}
    \centering
    \includegraphics[width=\linewidth]{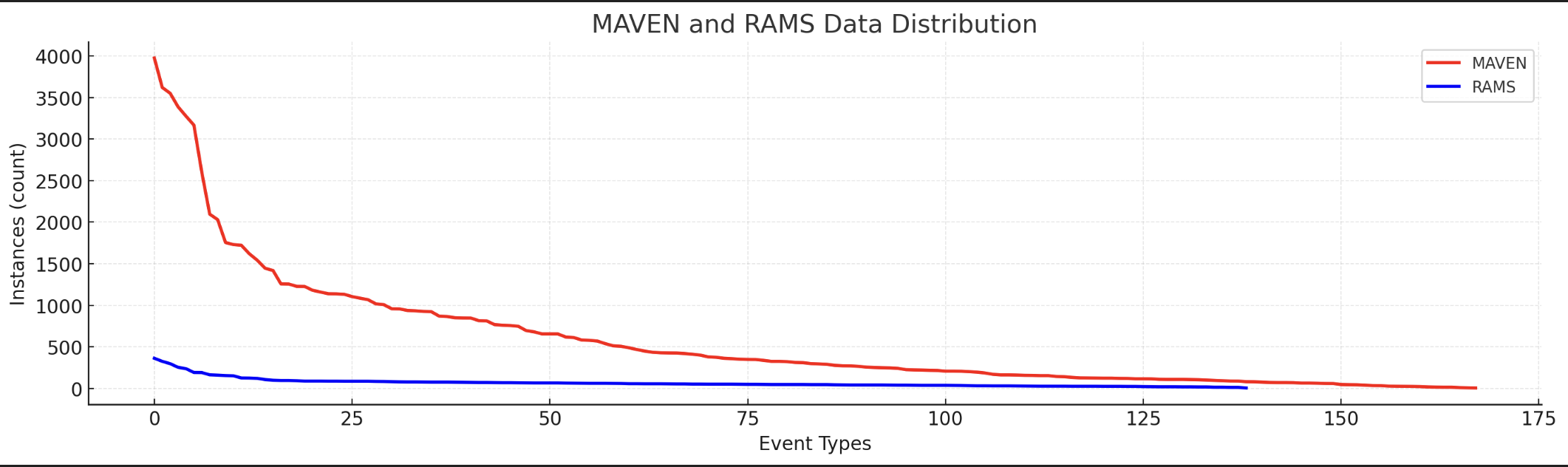}
    \caption{Distribution of event types by event mention instances in the MAVEN and RAMS dataset.}
    \label{fig:event-mention-distribution}
\end{figure}

\subsection{Task Definition}
\label{ssec:task_definition}

We follow the event detection task definition of the MAVEN dataset~\cite{wang-etal-2020-maven} which splits the problem into two parts \textbf{trigger identification (TI)}---the task of annotating which word spans in the text express an event in the sentence---and \textbf{trigger classification (TC)}---the task of classifying the event type that is expressed. Formally, in a sentence $s$ of $n$ words, $s = [w_1, w_2, ..., w_n]$, which expresses $m$ events, $e_j \in E$ where $|E| = m$, the events are defined by the triple $e_j = (f_{e_j}, l_{e_j}, t_{e_j})$ where $[w_{f_{e_j}}, w_{{l_{e_j}} - 1}]$ is the span of text that expresses event $e_j$ and $t_{e_j}$ is its event type. TI is then the task of extracting out a set of first and last word index pairs $(f, l)$ and TC is the task of extracting out the set of triples $(f, l, t)$ by adding in the type. Both tasks are evaluated using F1 scores. We focus on TC as TI is only an intermediate step for accomplishing TC.


\subsection{Prompted LLMs}
\label{ssec:prompted-llms}
\begin{figure}[ht]
\vspace{-2mm}
\centering
\begin{scriptsize}

\begin{tcolorbox}[colback=blue!9!white]

\begin{verbatim}
# Few-shot Prompt
input_sentence = join_tokens([[sentence_tokens]])

If a sentence contains multiple triggers, each trigger 
should be output separately as an independent JSON entry.
Ignore any errors in the response and extract only the 
relevant information needed for the JSON output.

prompt = " ".join([[few_shot_example]])

Identify the trigger word(s) and event type(s) from
the list in the sentence below.
The output must be strictly formatted as follows:
'[{{"sentence": ["word1", "word2", ...],
   "trigger": <trigger_text>,
   "e_start": <start_index>,
   "eventtype": <event_type>}}]'
\end{verbatim}
\end{tcolorbox}

\caption{Few-shot Prompt: Uses example-based contextual learning 
with explicit JSON formatting and multi-trigger handling.}

\vspace{2mm}

\begin{tcolorbox}[colback=green!9!white]
\begin{verbatim}
# Zero-shot Prompt
For sentence: [[tokenized_sentence]], identify all words or 
phrases that trigger an event. 

Each trigger must be one or two words long. Two-word triggers 
must be enclosed in the same double quotes.

Do not add comments, disclaimers, or examples before or 
after the JSON output. Do not use any symbols other than 
those explicitly required.
If a sentence contains multiple triggers, each trigger 
should be output separately as an independent JSON entry.
Ignore any errors in the response and extract only the 
relevant information needed for the JSON output.

The output must be strictly formatted as follows:
'[{{"sentence": ["word1", "word2", ...],
   "trigger": <trigger_text>,
   "e_start": <start_index>,
   "eventtype": <event_type>}}]'
\end{verbatim}
\end{tcolorbox}
\caption{Zero-shot Prompt: Relies purely on explicit 
instruction-based reasoning with structured JSON output.}

\end{scriptsize}

\vspace{-2mm}

\label{fig:combined-prompt-single}
\vspace{-3mm}
\end{figure}
We categorize model designs into two broad categories. One where LLMs are only prompted using in-context learning and another where
an LLM is used as a text embedding model in a wider neural architecture.

We additionally explore the use of LLMs through prompting, leveraging their in-context learning capabilities~\cite{brown-etal-2020-language}. This approach bypasses the need for computationally intensive model training by querying the model directly, offering a lightweight method for generalization.

We evaluate both few-Shot Prompt and zero-shot prompting. All evaluations are performed without updating model parameters. In the few-shot setting, we supply natural language demonstrations with labeled examples, each consisting of a sentence, a trigger word, and an event type. We did not provide the event descriptions, as there can be some datasets without publicly available annotation guidelines. Therefore, we tried to experiment without such descriptions. The experiments were conducted with varying temperature and number of shots~(\Cref{app:hyperprompt}).

\subsection{LLM-as-Embedding}
\label{ssec:llm-as-embedding}

\begin{figure}[t!]
    \centering
    \includegraphics[width=\linewidth]{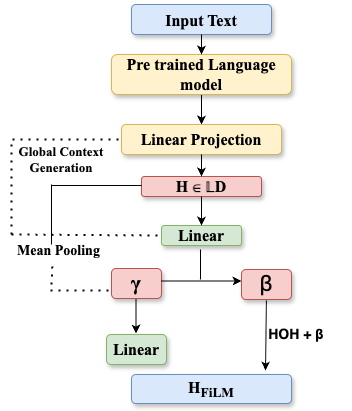}
    \caption{Overview of the Event Detection Framework}
    \label{fig:intro-figure}
\end{figure}

We use the base model from \citeposs{wang-etal-2023-continual} experiments. This is a relatively simple but effective neural model using text embeddings (Figure \ref{fig:intro-figure}). We use LLMs to generate embeddings for each word in the sentence which are then processed through a linear layer using dropout~\cite{srivastava-etal-2014-dropout} and layer normalization~\cite{ba2016layernormalization} followed by a linear softmax classifier. The TC task is then converted to a word-level classification task where each word is classified into an event type or ``NA'' if it is not an event trigger. Outputs for TI are computed by merging all valid event types into a single ``trigger'' classification label. This is a supervised model where the classification layers are trained concurrently with the finetuning of the LLM providing the embeddings. We do not use the data augmentation, knowledge transfer, or pivotal knowledge distillation techniques in our experiments, as these techniques were primarily used to improve continual learning.

\begin{figure*}
    \centering
    \includegraphics[width=\linewidth]{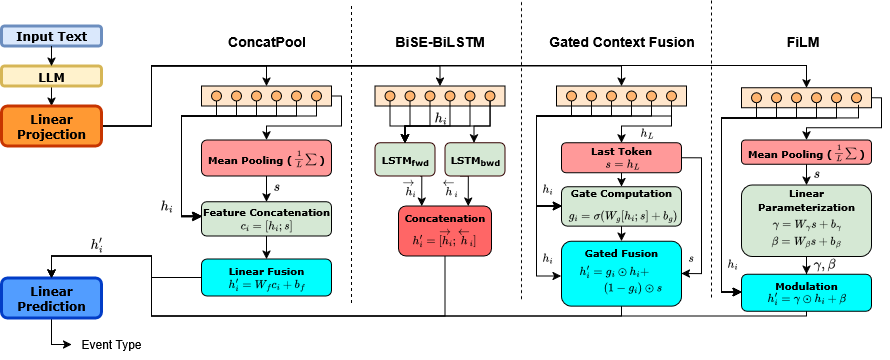}
    \caption{Diagrams for the four non-baseline LLM-as-Embedding variations to introduce sentence-level information to decoder-only models.}
    \label{fig:decoder-variations}
\end{figure*}

Following are the variations for incorporating 
sentence-level information 
that we tested for the decoder-only models. The BaseTE model neither introduces additional layers to the base model nor incorporates sentence-level context into the token representations. In contrast, the other architectures apply at least one of these modifications. All variations in this section have access to the LLM final layer hidden states, $H \in \mathbb{R}^{L \times D_{\text{LLM}}}$, where $L$ is the sequence length and $D_{\text{LLM}}$ is the LLM's hidden dimension. The variations are illustrated in ~\Cref{fig:decoder-variations}.

\subsubsection{BaseTE: Baseline Trigger Encoder}
%
%

Our baseline Trigger Encoder~(TE) linearly projects final layer hidden states from LLMs into the target dimensions with dropout ($p=0.2$) for regularization, GELU~\citep{hendrycks2016gelu} activation, and layer normalization~\citep{ba2016layernormalization}.


\subsubsection{ConcatPool}
\label{sssec:concatpool}
%
%
%

ConcatPool performs mean pooling and concatenation to produce sentence-contextualized hidden states.

\paragraph{1. Global Context Generation.} First, a single sentence-level embedding, $s \in \mathbb{R}^{D}$, is produced by applying mean pooling across the sequence of initial $L$ token representations, $s = \frac{1}{L} \sum_{i=1}^{L} h_i$.
    
\paragraph{2. Feature Concatenation.} $s$ is concatenated with each token representation, $h_i \in H$ to produce $c_i = [h_i ; s]$, a combined feature vector for each token position where $c_i \in \mathbb{R}^{2D}$.
    
\paragraph{3. Linear Fusion.} Finally, a linear fusion layer, with a learned weight matrix $W_{f} \in \mathbb{R}^{D \times 2D}$ and bias vector $b_{f} \in \mathbb{R}^{D}$, maps these concatenated vectors back to the original hidden dimension $D$, producing $h'_{i} = W_{f} c_i + b_{f}$. Here, $h'_i$ is the final output representation.

\subsubsection{FiLM: Feature-wise Linear Modulation}
\label{sssec:film}

%
%
%
%

This variation is inspired by \citet{DBLP:journals/corr/abs-1709-07871}, where a similar technique is successful in visual question answering. The core of the architecture is the FiLM-based context modulation, which consists of three steps.

\paragraph{1. Global Context Generation.} Mean pooling-based sentence-level embedding production in the same manner as ConcatPool~(\Cref{sssec:concatpool}).

\paragraph{2. FiLM Parameter Generation.} This global context vector $s$ is then passed through two separate linear layers, which act as a FiLM generator, to produce a feature-wise scaling vector $\gamma$ and a shifting vector $\beta$.
    \begin{equation}
        \gamma = W_{\gamma}s + b_{\gamma} \quad \text{and} \quad \beta = W_{\beta}s + b_{\beta}
        \label{eq:film_parameters}
    \end{equation}

\paragraph{Feature-wise Modulation.} Finally, the generated $\gamma$ and $\beta$ parameters are used to modulate the initial token representations $H$ via an element-wise affine transformation.
    \begin{equation}
        H_{\text{FiLM}} = \gamma \odot H + \beta
        \label{eq:film_modulation}
    \end{equation}

\subsubsection{BiSE‑BiLSTM: Bidirectional Sequential Encoding using BiLSTM}
%
%
%
%

This variation uses a bidirectional LSTM~\cite{huang2015bidirectionallstmcrfmodelssequence} over the sequence of transformer hidden states to produce new bidirectionally contextualized tokens. For each token $h_i$ in the sequence $H$, the forward and backward LSTMs produce the forward $\overrightarrow{h_i}$ and backward $\overleftarrow{h_i}$ hidden states, respectively, using the following recursive computations: $\overrightarrow{h_i} = \text{LSTM}_{\text{fwd}}(h_i, \overrightarrow{h_{i-1}})$, $\overleftarrow{h_i} = \text{LSTM}_{\text{bwd}}(h_i, \overleftarrow{h_{i+1}})$. The final representation for each token, $h'_i$, is formed by concatenating the hidden states from both directions, $h'_i = [\overrightarrow{h_i} ; \overleftarrow{h_i}]$.


\subsubsection{Gated Context Fusion}

%
%
%

This variation fuses token- and sequence-level information using a gating function through these three stages.

\paragraph{1.\ Global Context Generation.} 
    A single sequence-level embedding, $s \in \mathbb{R}^{D}$, is acquired by taking the final layer hidden state of the last token in the sequence.
    
\paragraph{2. Gate Computation.} A fusion gate, $g_i \in [0, 1]^D$, is computed for each token representation $h_i \in H$. The concatenation of the token, $h_i$, and sequence, $s$, embeddings are passed through a linear layer ($W_g, b_g$) followed by a sigmoid activation function, $\sigma$.
    \begin{equation}
        g_i = \sigma(W_g [h_i ; s] + b_g)
        \label{eq:gate_computation}
    \end{equation}
    The gate $g_i$ acts as an adaptive switch, determining the proportion of information to retain from each source in the next step.
    
\paragraph{3. Dynamic Fusion.} The final representation for each token, $h'_i$, is produced by performing a $g_i$-controlled weighted sum. This interpolates between the original token representation $h_i$ and the sequence vector $s$.
    \begin{equation}
        h'_i = g_i \odot h_i + (1 - g_i) \odot s
        \label{eq:gated_fusion}
    \end{equation}
    where $\odot$ denotes element-wise multiplication.







%% file: sections/experimental_setup.tex
\subsection{Datasets}
The chosen datasets are sentence level ED datasets which have large number of classes, as our experiments will be on long-tailed classes. We used 2 datasets: MAVEN~\cite{wang-etal-2020-maven} and RAMS~\cite{ebner-etal-2020-multi}. MAVEN is a dataset of 4,480 Wikipedia documents, 118,732 event mentions, and 168 event types. The RAMS dataset comprises 139 event types and 9,124 event mentions from news articles.


\subsection{Evaluation Metrics}

We use Micro- and Macro-F1 trigger classification scores along with recall and precision. 
A prediction is labeled correct if and only if the word span and event type all match the gold labels. Micro-F1 is the standard metric used in prior work, where each event mention is weighted equally. We add Macro-F1 to our analysis, which is computed by taking the mean of the F1 score for each event type. This is a useful measure of generalizability because event type frequencies follow a long-tail distribution. \Cref{fig:event-mention-distribution} shows the distribution of event type instances across the entire MAVEN and RAMS datasets. The MAVEN has a larger imbalance among the classes.


\subsection{LLM Models}

For LLM-as-embedding models, we consider BERT~\cite{devlin2019bert}—specifically the bert-base-uncased model from Hugging Face~\cite{wolf2020transformers}—as well as Llama 3.2 1B, Qwen2 1.5B, bert-base-uncased, bert-large-uncased, RoBERTa-base, RoBERTa-large~\cite{liu2019roberta}, T5-base, T5-large~\cite{raffel2020exploring}, Llama 3.2 3B, Qwen2.5 3B, DeBERTa-base, and DeBERTa-large~\cite{he2021deberta}. The largest possible Llama models were selected based on the computational resources available.

For few-shot and zero-shot prompting experiments, we used Llama 3 8B~\cite{touvron2024llama}, Qwen2.5 7B~\cite{qwen2025qwen25technicalreport}, Gemma 8B~\cite{gemmateam2024gemmaopenmodelsbased}, and DeepSeek 7B~\cite{deepseekai2024deepseekllmscalingopensource}.


\subsection{Baselines}

We compare our model variations against fully finetuned models, zero-shot and few-shot results.


\subsection{Hyperparameter Selection}
The LLM-as-embedding model with BERT uses the best hyperparameters from \citet{wang-etal-2023-continual}. A short hyperparameter search of the learning rate was performed for the other models. The exact default values for all models are provided in \Cref{app:default_hyperparameters}.

Wherever LoRA is used in our experiments, it is applied to all the linear layers of the model.



%% file: sections/results.tex
\begin{table*}[ht]
\centering
\resizebox{\textwidth}{!}
{
\begin{tabular}{|c|c|c|c|c|c|c|c|c|c|c|c|}
\hline
\textbf{Dataset} & \textbf{Model} & \textbf{Size} & \textbf{Temp} & \textbf{Shots} & \textbf{Min Count} & \textbf{Shot Type} & \textbf{Macro-P(\%)} & \textbf{Macro-R(\%)} & \textbf{Micro-F1(\%)} & \textbf{Macro-F1(\%)} \\
\hline
\multirow{2}{*}{RAMS} & Llama 3 & 8B & 0.4 & 3\% & 3 & Few-shot & 26.25 & 29.87 & 34.23 & 27.94 \\
&Qwen 2.5 & 7B & 0.4 & 3\%& 3 & Few-shot & 26.14 & 30.52& 33.32 & 28.16\\
\hline

\multirow{7}{*}{MAVEN} & Llama 3 & 8B & 0.4 & 3\% & 3 & Few-shot & \textbf{56.82} & \textbf{51.72} & \textbf{63.95} & \textbf{54.13} \\
& Qwen 2.5 & 7B & 0.4 & 3\% & 3 & Few-shot & 55.51 & 50.68 & 61.87 & 52.99 \\
& Gemma & 8B & 0.4 & 3\% & 1 & Few-shot & 54.02 & 49.21 & 61.31 & 51.49 \\
 & DeepSeek & 7B & 0.4 & 1\% & 3 & Few-shot & 53.13 & 48.14 & 61.01 & 50.53 \\
 & Llama 3 & 8B & 0.4 & 0\% & 0 & Zero-shot & 51.36 & 46.19 & 36.10 & 24.11  \\
 & Qwen 2.5 & 7B & 0.4 & 0\% & 0 & Zero-shot & 50.11 & 45.15 & 33.23 & 22.75 \\
 & Gemma & 8B & 0.4 & 0\% & 0 & Zero-shot & 51.24 & 46.29 & 34.76 & 22.95 \\
\hline
\end{tabular}
}
\caption{Best-performing Few-shot and Zero-shot Configurations}
\label{tab:best_macro_f1_with_dataset_llama}
\end{table*}

\begin{table*}[ht]
\centering
\scriptsize
\resizebox{\textwidth}{!}{
\begin{tabular}{|l|l|c|c|c|c|c|c|}
\hline
\textbf{Model} & \textbf{Variation} & \textbf{Size} & \textbf{LoRA} & \textbf{Macro-P(\%)} & \textbf{Macro-R(\%)} & \textbf{Micro-F1(\%)} & \textbf{Macro-F1(\%)} \\
\hline
\multirow{12}{*}{Llama} 
 & BaseTE & \multirow{7}{*}{1B} & Yes & 64.02 & 58.54 & 67.79 & 59.31 \\
 & Gating &  & Yes & 63.54 & 60.11 & 68.26 & 59.94 \\
 & FiLM &  & Yes & 62.44 & 64.14 & 68.13 & \textbf{61.38} \\
 & BiSE-BiLSTM &  & Yes & 63.33 & 62.31 & 67.76 & 60.64 \\
 & ConcatPool &  & Yes & 59.40 & 65.14 & 67.70 & 60.56 \\
 & (None) &  & No & 61.32 & 55.84 & 64.37 & 57.23 \\
 
\cline{2-8}
 & BaseTE & \multirow{3}{*}{3B} & Yes & 65.71 & 61.14 & 68.61 & 61.63 \\
 & Gating &  & Yes & 65.96 & 59.60 & 67.90 & 60.59 \\
 & FiLM &  & Yes & 60.76 & \textbf{65.97} & 68.13 & \textbf{62.07} \\
\hline
\multirow{11}{*}{Qwen} 
 & BaseTE & \multirow{7}{*}{1.5B} & Yes & 63.63 & 59.17 & 68.11 & 59.37 \\
 & Gating &  & Yes & 65.41 & 58.29 & 67.16 & 59.63 \\
 & FiLM &  & Yes & 59.25 & \textbf{65.86} & 67.86 & 60.61 \\
 & BiSE-BiLSTM &  & Yes & 65.18 & 61.47 & 67.76 & \textbf{61.20} \\
 & ConcatPool &  & Yes & 60.79 & 64.41 & 67.64 & 60.85 \\
 & (None) &  & No & 62.03 & 57.03 & 65.89 & 57.21 \\
\cline{2-8}
 & BaseTE & \multirow{4}{*}{3B} & Yes & 65.49 & 61.56 & \textbf{68.77} & 61.30 \\
 & FiLM &  & Yes & 62.52 & 64.34 & 68.05 & \textbf{62.01} \\
 & Gating &  & Yes & 64.62 & 60.83 & 67.91 & 60.48 \\
 & BiSE-BiLSTM &  & Yes & 63.44 & 60.31 & 67.71 & 60.20 \\
\hline
Dec2Enc Llama & & 1B & Yes & 65.36 & 58.47 & 68.24 & 59.74 \\
\hline
LLM2Vec Llama & & 1B & Yes & 61.33 & 55.23 & 66.1 & 56.49 \\
\hline
\multirow{4}{*}{BERT} & \_ & base & No & 58.31 & 59.33 & 67.61 & 56.05 \\
 & & base & Yes & 63.65 & 55.36 & 67.26 & 57.26 \\
 & & large & No & 64.51 & 58.81 & 67.48 & \textbf{59.99} \\
 & & large & Yes & 64.72 & 55.03 & 67.52 & 57.45 \\
\hline
\multirow{4}{*}{T5} & \_ & large & No & 63.10 & 56.56 & 66.10 & \textbf{57.30} \\
 & & large & Yes & \textbf{66.70} & 44.73 & 61.93 & 51.03 \\
 & & small & No & 60.92 & 50.32 & 64.25 & 52.71 \\
 & & small & Yes & 65.11 & 50.53 & 64.45 & 54.16 \\
\hline
\multirow{4}{*}{Roberta} & \_ & base & No & 63.18 & 62.48 & \textbf{68.93} & 61.41 \\
 & & large & No & 64.49 & 60.23 & 68.01 & 60.53 \\
 & & base & Yes & 64.77 & 63.07 & 68.83 & \textbf{61.50} \\
 & & large & Yes & 61.56 & 63.46 & 67.61 & 60.86 \\
\hline
\multirow{4}{*}{DeBERTa-v3} & \_ & base & No & 62.39 & 60.45 & 66.78 & \textbf{60.43} \\
 & & base & Yes & 61.28 & 61.68 & 67.93 & 59.35 \\
 & & large & No & 59.75 & 62.69 & 68.19 & 59.76 \\
 & & large & Yes & 61.99 & 61.2 & \textbf{68.37} & 60.01 \\
\hline
\end{tabular}
}
\caption{Evaluation metrics for MAVEN dataset}
\label{tab:other_metrics}
\end{table*}

\begin{table*}[ht]
\centering
\resizebox{0.9\textwidth}{!}{
\begin{tabular}{|l|c|c|c|c|c|c|c|c|c|}
\hline
\textbf{Model} & \textbf{Variation} & \textbf{Size} & \textbf{LoRA} & \textbf{Micro-P(\%)} & \textbf{Micro-R(\%)} & \textbf{Macro-P(\%)} & \textbf{Macro-R(\%)} & \textbf{Micro-F1(\%)} & \textbf{Macro-F1(\%)} \\
\hline
Llama & (None) & 1B & no & 35.01 & 31.48 & 32.20 & 27.89 & 33.15 & 33.19\\ 
Llama & FiLM & 1B & yes & 37.43 & 34.30 & 34.73 & 28.48 & 35.78 & 28.28  \\ 
Qwen & (None) & 1.5B & no & 36.39 & 32.68 & 32.88 & 24.64 & 34.43 & 25.82\\ 
Qwen & FiLM & 1.5B & yes & 38.58 & 37.37 & 35.76 & 31.38 & 37.97 & 30.25 \\ 
BERT & (None) & small & no & 37.66 & 36.40 & 34.96 & 30.39 & 37.02 & 30.74\\ 
RoBERTa & (None) & small & no & 41.45 & 41.45 & 37.05 & 33.25 & 41.45 & 30.85\\ 
T5 & (None) & small & no & 26.52 & 38.63 & 25.00 & 30.50 & 31.45 & 24.04 \\ 
\hline
\end{tabular}
}
\caption{Evaluation metrics for RAMS dataset}
\label{tab:rams main metrics}
\end{table*}

\begin{table}[ht]
\centering
\resizebox{\columnwidth}{!}
{
\begin{tabular}{|l|l|l|l|l|}
    \hline
    \textbf{Model} & \textbf{Dropout} & \textbf{Size} & \textbf{Micro-F1(\%)} & \textbf{Macro-F1(\%)} \\
    \hline
    & 0(Default) & 1B & 64.37 & 57.23 \\
    \hline
    Llama (full) & 0.3 & 1B & 65.89 & 57.22 \\
    \hline
     & 0(Default) & 1.5B & 65.89 & 57.21 \\
    Qwen (full) & 0.3 & 1.5B & 65.05 & 57.68 \\
    \hline
\end{tabular}
}
\caption{Performance with varying dropout rates in after finetuning the whole models}
\label{tab:dropout tests}
\end{table}




\subsection{Supervised Finetuning Results}
\paragraph{LoRA enhances Micro-F1 score of the decoder-only models.} \Cref{tab:other_metrics} illustrates that finetuning the decoder-only models (Llama 1B, Qwen 1.5B) alone did not surpass the smaller BERT or RoBERTa models in terms of Macro-F1 or Micro-F1 scores. With LoRA, however, the decoder-only models improved substantially, surpassing finetuned BERT models in F1 scores. We find that increasing the attention dropout of these models alone did not lead to the same performance gains \Cref{tab:dropout tests}, indicating that LoRA itself provides a boost in performance that is separate from typical regularization in addition to improving computational efficiency.

\paragraph{LoRA improves long-tailed class performance.} LoRA increased the Macro-F1 scores for most of the models, especially all of the Llama and Qwen models, indicating that LoRA effectively handled long-tailed classes. How evaluation metrics change compared to frequency of classes is shown in~\Cref{app:metrics-distribution}. Scaling up model size tends to improve the performance, albeit with computational cost(~\Cref{tab:other_metrics}).



\begin{figure}[ht]
    \centering
    \includegraphics[width=\linewidth]{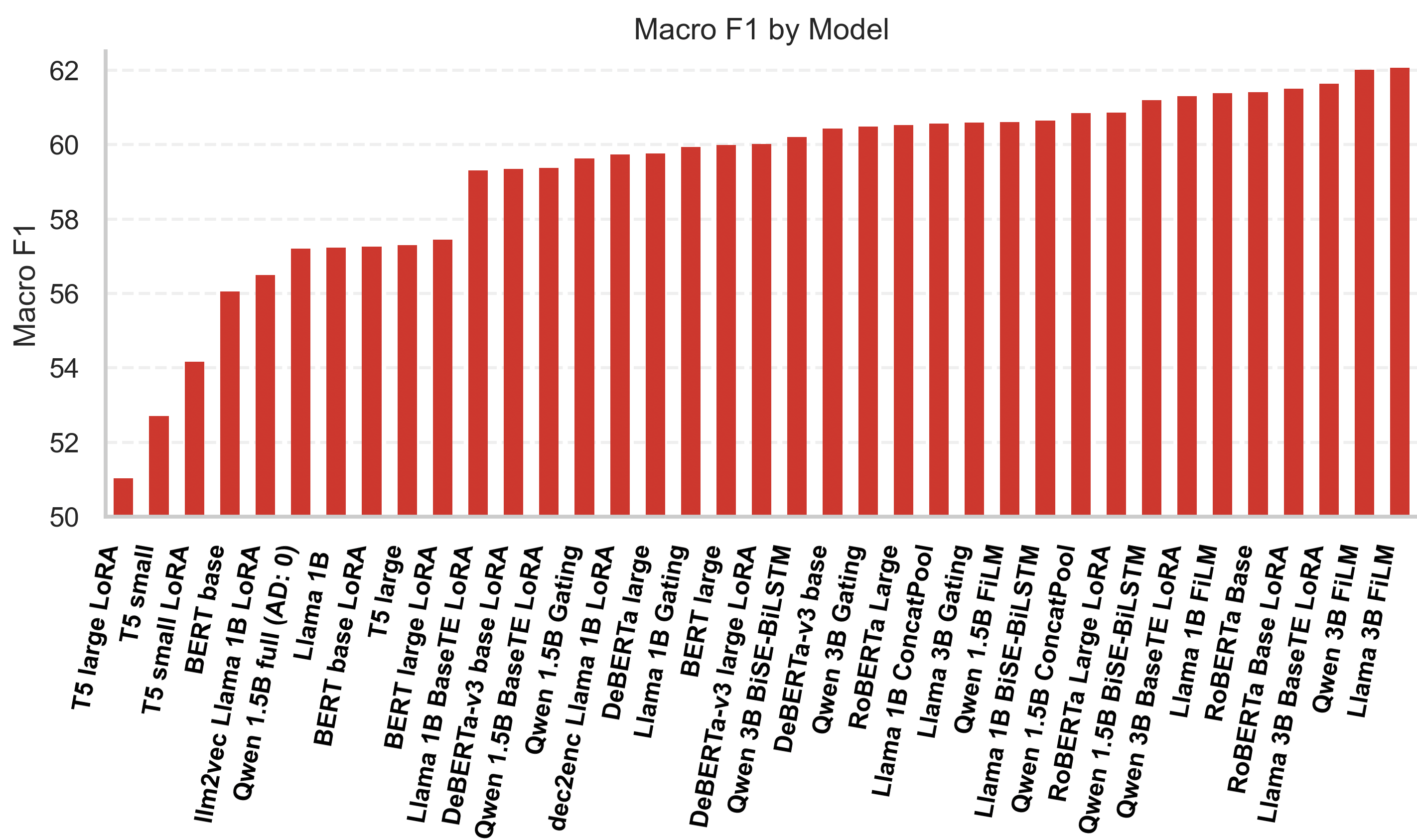}
    \caption{Macro-F1 performance of all finetuned models on MAVEN}
    \label{fig:macro-f1-finetuned-models}
\end{figure}

\paragraph{Sentence-level context improves Llama and Qwen models, especially in terms of Macro-F1 scores.} FiLM with mean pooling and BiSE-BiLSTM was the best overall across the models(~\Cref{fig:macro-f1-finetuned-models}). Qwen 3B model with FiLM variation had the best Macro-F1 score (62.01\%) for all experimental settings. For Qwen 1.5B, even the BiSE-BiLSTM variant was competitive with a Macro-F1 of 61.20\%. For Llama 1B model, FiLM had the best Macro-F1 score (61.38\%). Almost all the context variations improved the decoder-only models.

Dec2Enc (Llama 1B) with LoRA outperforms the BaseTE Llama with LoRA model. But the FiLM along with some other variations of Llama turned out to have superior performance in terms of Macro-F1 scores. However, LLM2Vec (Llama 1B) with LoRA did not show any meaningful gain.

\begin{figure}[ht]
    \centering
    \includegraphics[width=\linewidth]{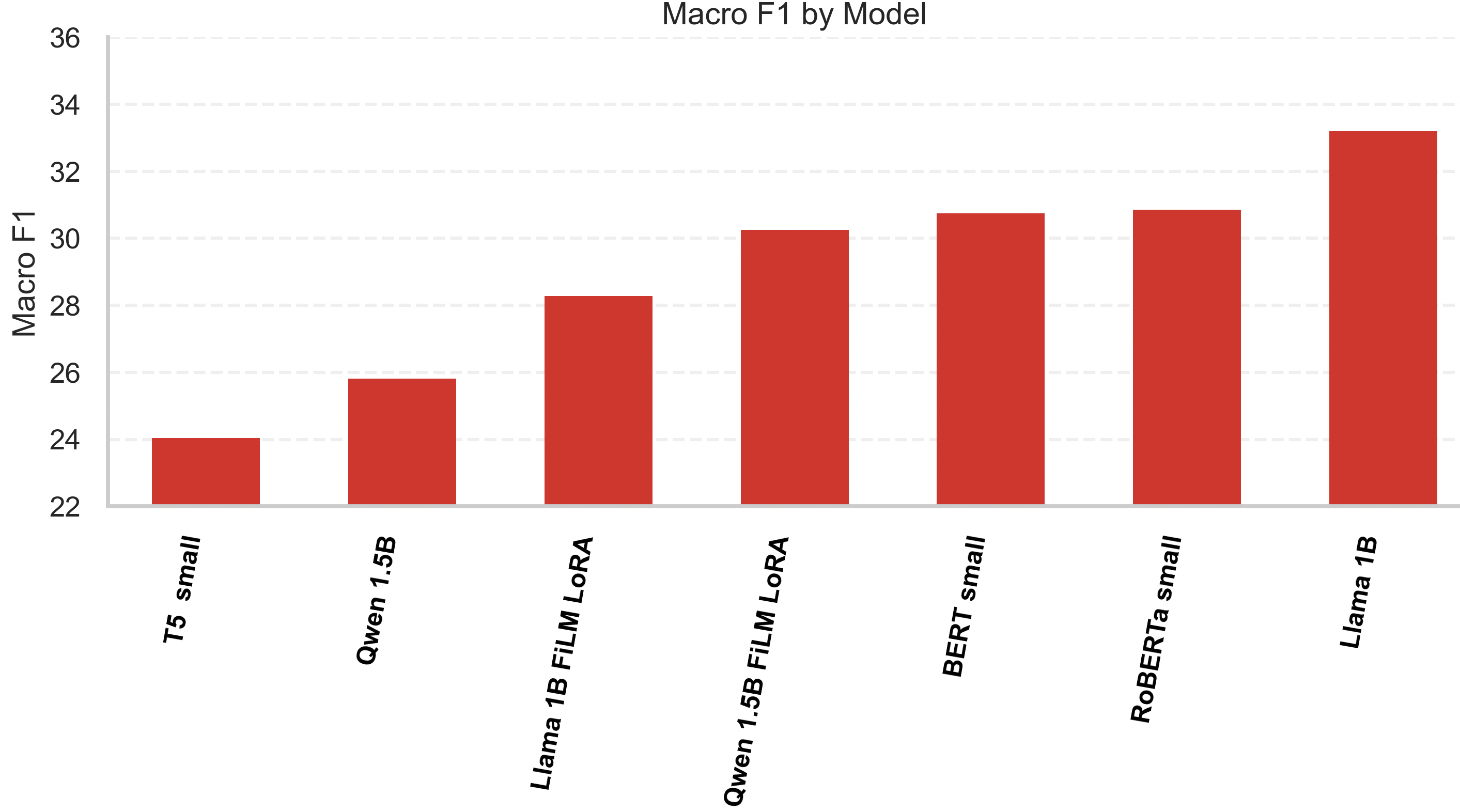}
    \caption{Macro-F1 performance of finetuned models on RAMS}
    \label{fig:macro-f1-finetuned-models_rams}
\end{figure}


\paragraph{The decoder-only models, especially with those variations, showed a tendency to hold a high recall score.} So, for tasks where high recall score is paramount, these models can be a good choice.


\textbf{The T5 models have a unique ability of holding high precision scores.} If any task requires conservative prediction of events, then T5 is a good candidate. The DeBERTa-v3 large model with LoRA finetuning had the best Macro-Recall (67.47\%) among all the models. The Micro-F1 score, Precision, and Recall are illustrated in ~\ref{app:micro-f1}~\ref{app:precision}~\ref{app:recall}.

\textbf{Experiments on the RAMS dataset showed similar pattern in the results.} The only notable exception is that the full Llama finetuning has a better Macro-F1 that the Llama-FiLM variation(~\Cref{tab:rams main metrics}). However, the Macro-F1 score for the FiLM variations are better than the fully finetuned decoder-only models (\Cref{fig:macro-f1-finetuned-models_rams}). Also, in this case, BERT and RoBERTa showed slightly better Macro-F1 scores than Llama and Qwen variations. Qwen FiLM still has the second best Micro-F1 score after RoBERTa. 

However, the difference between the few-shot results and finetuning results for RAMS are not much different. This indicates that for smaller datasets, finetuning might not boost the performance of these models compared to few-shot setting.

\subsection{Results of Prompt-based Techniques}

We evaluated the performance of several instruction-tuned large language models (LLMs) on the MAVEN event detection dataset.
\subsubsection{Zero-Shot Performance}

As expected, zero-shot prompting yielded substantially lower performance, especially in Macro-F1, highlighting the difficulty LLMs face when generalizing to tail classes without in-context examples. Llama 3 (8B) achieved the best zero-shot Micro-F1 of 36.10\%, but the Macro-F1 dropped to 24.11\%. Qwen 2.5 (7B) and Gemma (8B) showed similar trends, with Macro-F1 values of 22.75\% and 22.95\%, respectively.

\subsubsection{Few-Shot Performance}
Few-shot prompting consistently outperformed zero-shot prompting across all models in both Micro-F1 and Macro-F1 metrics, underscoring the effectiveness of in-context learning for long-tailed event detection. Among the models, Llama 3 (8B) achieved the highest performance on the MAVEN dataset, with its best configuration---3\% shots, min count of 3, and temperature 0.4---yielding a Micro-F1 of 63.95\% and Macro-F1 of 54.13\%. 

On the RAMS dataset, the same configuration of Llama 3 (8B) achieved a Micro-F1 of 34.23\% and Macro-F1 of 27.94\%, while Qwen 2.5 (7B) obtained a Micro-F1 of 33.32\% and Macro-F1 of 28.16\% . These results indicate that few-shot learning significantly enhances generalization across both large-scale and domain-specific event extraction benchmarks. The few-shot results hallucinated by introducing new event types, which is discussed in ~\Cref{app:hallucination}.

\subsubsection{Statistical Significance Testing}
We conducted a paired t-test ~\cite{Virtanen2020SciPy} on the Macro-F1 scores to determine whether our findings differ significantly from each other. The results showed that the Variations of Llama and Qwen models are statistically significantly different (p<0.05) than the fully finetuned models, BERT, and T5 models. But, some of their F1 scores are not significantly different than the RoBERTa model predictions (\Cref{fig:heatmap}).

%% file: sections/conclusion.tex
Introducing bidirectional attention or sentence context boosted the effectiveness of the decoder-only models. It enabled Llama and Qwen surpass the BERT models and made them at least as effective as RoBERTa models. This shows that the decoder-only models can be powerful even for ED tasks when necessary adaptations are made.

Additionally, different models exhibit varying strengths in terms of precision and recall. Our study offers practical guidance for selecting models based on the specific needs of an application, whether it prioritizes precision or recall.


Further research on how sentence length affects these models will give us better insight. As most of the recent LLMs are generative in nature and built upon decoder-only architectures, exploring their adaptation for event detection remains a crucial and promising direction for future research. These adaptations can further help develop more robust NLU systems as well.

%% file: appendices/default_hyperparameters.tex
\appendix
\section{Additional evaluation Metrics}
\label{app: additional metrics finetuning}
Here are Micro-Precision and Micro-Recall scores of our fine-tuned models in table ~\ref{tab:micro_metrics}
\begin{table*}[ht]
\centering
\scriptsize
\resizebox{0.7\textwidth}{!}{
\begin{tabular}{|l|l|c|c|c|c|}
\hline
\textbf{Model} & \textbf{Variation} & \textbf{Size} & \textbf{LoRA} & \textbf{Micro-P(\%)} & \textbf{Micro-R(\%)} \\
\hline
\multirow{12}{*}{Llama} & BaseTE & 1B & Yes & 68.47 & 67.13 \\
 & Gating & 1B & Yes & 68.58 & 67.94 \\
 & FiLM & 1B & Yes & 65.80 & 70.63 \\
 & BiSE-BiLSTM & 1B & Yes & 65.26 & 70.45 \\
 & ConcatPool & 1B & Yes & 63.73 & 72.20 \\
 & BaseTE & 3B & Yes & 68.81 & 68.40 \\
 & Gating & 3B & Yes & 69.68 & 66.22 \\
 & FiLM & 3B & Yes & 65.15 & 71.41 \\
 & Dropout: 0 & 1B & No & 65.35 & 63.42 \\
 & Dropout: 0.3 & 1B & No & 66.54 & 65.25 \\
\hline
DEC2ENC Llama & & 1B & Yes & 70.19 & 66.39 \\
\hline
LLM2VEC Llama & & 1B & Yes & 67.85 & 64.44 \\
\hline
\multirow{4}{*}{BERT} & \_ & base & No & 64.51 & 71.03 \\
 & & base & Yes & 68.91 & 63.36 \\
 & & large & No & 68.51 & 66.49 \\
 & & large & Yes & 69.26 & 65.88 \\
\hline
\multirow{4}{*}{T5} & \_ & large & No & 67.10 & 65.14 \\
 & & large & Yes & 74.44 & 53.03 \\
 & & small & No & 67.93 & 60.96 \\
 & & small & Yes & 70.00 & 59.73 \\
\hline
\multirow{4}{*}{Roberta} & \_ & base & No & 67.74 & 70.19 \\
 & & large & No & 68.45 & 67.59 \\
 & & base & Yes & 67.28 & 70.46 \\
 & & large & Yes & 64.10 & 71.53 \\
\hline
\multirow{11}{*}{Qwen} & BaseTE & 1.5B & Yes & 67.22 & 69.03 \\
 & Gating & 1.5B & Yes & 70.25 & 64.33 \\
 & FiLM & 1.5B & Yes & 63.55 & 72.81 \\
 & BiSE-BiLSTM & 1.5B & Yes & 67.05 & 68.47 \\
 & ConcatPool & 1.5B & Yes & 65.11 & 70.37 \\
 & Dropout: 0 & 1.5B & No & 67.77 & 64.12 \\
 & Dropout: 0.3 & 1.5B & No & 69.09 & 61.45 \\
 & BaseTE & 3B & Yes & 67.71 & 69.87 \\
 & FiLM & 3B & Yes & 65.53 & 70.79 \\
 & Gating & 3B & Yes & 69.02 & 66.84 \\
 & BiSE-BiLSTM & 3B & Yes & 67.58 & 67.85 \\
\hline
\multirow{3}{*}{DeBERTa-v3} & \_ & base & No & 66.71 & 66.87 \\
 & & base & Yes & 66.13 & 69.84 \\
  & & large & No & 67.02 & 69.42 \\
 & & large & Yes & 67.47 & 69.3 \\
\hline
\end{tabular}
}
\caption{Micro-Precision and Micro-Recall}
\label{tab:micro_metrics}
\end{table*}

\begin{table*}[ht]
\centering
\resizebox{\textwidth}{!}{
\begin{tabular}{|l|c|c|c|c|c|c|c|c|c|c|c|}
\hline
\textbf{Model} & \textbf{Size} & \textbf{Temp} & \textbf{Shots} & \textbf{Min Count} & \textbf{Shot Type} & \textbf{Micro-P(\%)} & \textbf{Micro-R(\%)} & \textbf{Micro-F1(\%)} & \textbf{Macro-P(\%)} & \textbf{Macro-R(\%)} & \textbf{Macro-F1(\%)} \\
\hline
\multirow{6}{*}{LLaMA 3} & \multirow{6}{*}{8B} & 0.0 & 1\% & 1 & \multirow{6}{*}{Few-shot} & 73.84 & 50.67 & 60.10 & 51.36 & 46.19 & 48.67 \\
&  & 0.4 & 1\% & 1 &  & 74.45 & 51.38 & 60.80 & 51.58 & 46.52 & 48.96 \\
 & & 0.4 & 1\% & 3 &  & 73.81 & 50.95 & 60.28 & 50.69 & 45.61 & 48.02 \\
 &  & 0.0 & 3\% & 3 &  & 75.06 & 51.90 & 61.37 & 54.17 & 48.61 & 51.28 \\
 & & 0.4 & 3\% & 3 &  & 77.94 & 54.21 & \textbf{63.95} & 56.82 & 51.72 & \textbf{54.13} \\
 &  & 0.4 & 3\% & 1 &  & 77.39 & 53.62 & 63.35 & 56.12 & 51.64 & 53.79 \\ \hline
\multirow{6}{*}{Qwen 2.5} & \multirow{6}{*}{7B} & 0.0 & 1\% & 1 & \multirow{6}{*}{Few-shot} & 74.29 & 50.24 & 59.94 & 50.11 & 45.15 & 47.50 \\
 &  & 0.4 & 1\% & 1 &  & 74.02 & 50.50 & 60.04 & 51.69 & 46.76 & 49.11 \\
 &  & 0.0 & 3\% & 3 &  & 74.42 & 51.42 & 60.82 & 52.64 & 47.38 & 49.87 \\
 &  & 0.4 & 3\% & 3 &  & 75.34 & 52.48 & 61.87 & 55.51 & 50.68 & 52.99 \\
 &  & 0.4 & 1\% & 3 &  & 74.35 & 50.37 & 60.05 & 50.23 & 45.82 & 47.94 \\
 &  & 0.4 & 3\% & 1 &  & 75.39 & 52.15 & 61.65 & 55.37 & 50.34 & 52.74 \\ \hline
\multirow{6}{*}{Gemma} & \multirow{6}{*}{8B} & 0.0 & 1\% & 1 & \multirow{6}{*}{Few-shot} & 74.75 & 51.16 & 60.78 & 51.24 & 46.29 & 48.64 \\
 &  & 0.4 & 1\% & 1 &  & 74.51 & 51.43 & 60.85 & 51.98 & 46.77 & 49.21 \\
 &  & 0.0 & 3\% & 3 &  & 74.18 & 51.61 & 60.87 & 51.96 & 47.13 & 49.42 \\
&  & 0.4 & 3\% & 3 &  & 75.40 & 52.66 & 62.01 & 54.03 & 48.81 & 51.29 \\
 &  & 0.4 & 1\% & 3 &  & 74.72 & 51.30 & 60.83 & 51.54 & 46.49 & 48.87 \\
 &  & 0.4 & 3\% & 1 &  & 74.33 & 52.17 & 61.31 & 54.02 & 49.21 & 51.49 \\ \hline
\multirow{2}{*}{DeepSeek} & \multirow{2}{*}{7B} & \multirow{2}{*}{0.4} & 1\% & 3 & \multirow{2}{*}{Few-shot} & 74.21 & 51.80 & 61.01 & 53.13 & 48.14 & 50.53 \\
 &  & & 3\% & 1 &  & 73.81 & 50.95 & 60.29 & 51.86 & 46.73 & 49.15 \\
\hline
LLaMA 3 & 8B & \multirow{3}{*}{0.4} & \multirow{3}{*}{0\% }& - & \multirow{3}{*}{Zero-shot} & 44.61 & 30.32 & 36.10 & 51.36 & 46.19 & 24.11 \\
Qwen 2.5 & 7B &  &  & - &  & 41.75 & 27.60 & 33.23 & 50.11 & 45.15 & 22.75 \\
Gemma & 8B &  &  & - &  & 43.51 & 28.94 & 34.76 & 51.24 & 46.29 & 22.95 \\
\hline
\end{tabular}
}
\caption{Few-shot and Zero-shot Event Detection Metrics}
\label{tab:all_metrics_complete}
\end{table*}

\section{Default Hyperparameters}
\label{app:default_hyperparameters}

The default hyperparameters of the LLM-as-embedding model (Section~\ref{ssec:llm-as-embedding}) are shown in Table~\ref{tab:default-hyperparameters}.

\begin{table*}[ht]
    \centering
    \begin{tabular}{l|l} 
        \hline
        \textbf{Hyperparameter} & \textbf{Value} \\
        \hline \hline
        Hidden dimension & 100 \\
        Last layer dropout & 0.2 \\
        Optimizer & AdamW \\
        Learning rate (LLaMA, Qwen, RoBERTa, T5, DeBERTa)\tablefootnote{For DeBERTa-large full fine-tuning, LR $=1 \times 10^{-6}$.} & $5 \times 10^{-6}$ \\
        Learning rate (BERT) & $5 \times 10^{-5}$ \\
        Warmup proportion & 0.1 \\
        Weight decay & 0.01 \\
        \hline
    \end{tabular}
    \caption{Default hyperparameters used for LLM-as-embedding models.}
    \label{tab:default-hyperparameters}
\end{table*}

\section{Hyperparameters for Prompting Techniques}
\label{app:hyperprompt}
To comprehensively assess few-shot performance under limited-resource settings, we experimented with four configurations: 1\% shots with 1-min count, 1\% shots with 3-min count, 3\% shots with 1-min count, and 3\% shots with 3-min count. These combinations were chosen to balance \textit{prompt space constraints} and \textit{event type coverage}. The 1\% settings mimic extremely low-resource conditions, while 3\% shots provide broader contextual diversity. Enforcing a minimum example count of 3 per event type ensures that even rare classes are represented, thereby testing model generalization to tail events. Conversely, a 1-min count allows for more event types to be included but may underrepresent rarer classes. This setup enables analysis of the trade-off between \textit{event diversity} and \textit{example richness per type}.

Moreover, a decoding temperature of 0.4 generally led to improved results compared to 0.0, likely due to increased output diversity and reduced repetition. These findings highlight the challenge of long-tailed event detection and demonstrate that even minimal prompting, when carefully structured, can substantially enhance performance.

\section{P values for t-test}
\begin{figure}[ht]
    \centering
    \includegraphics[width=\linewidth]{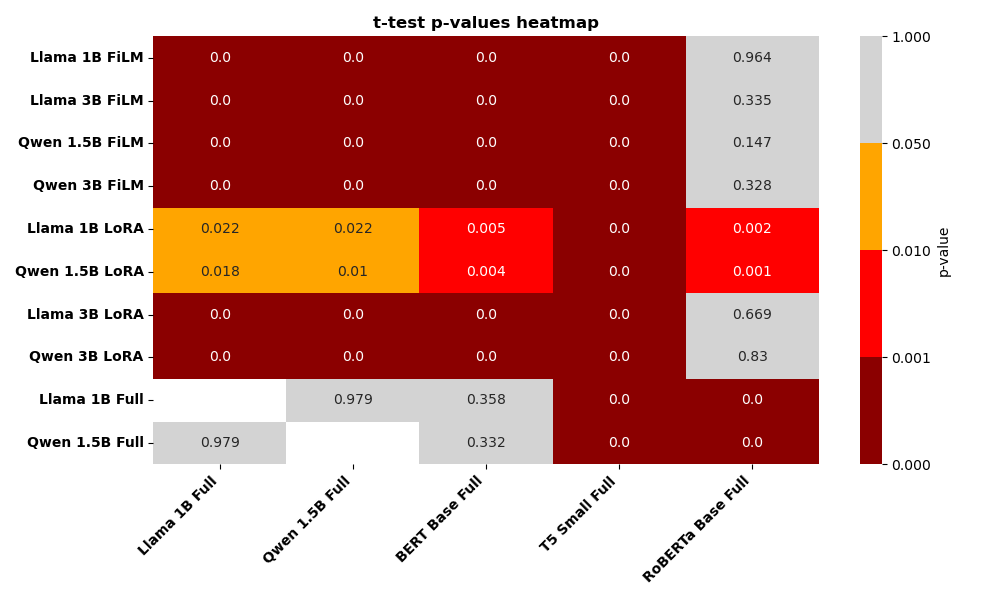}
    \caption{p values for t-test}
    \label{fig:heatmap}
\end{figure}

\section{Micro-F1 scores for fine-tuning results}
\label{app:micro-f1}

\begin{figure}[ht]
    \centering
    \includegraphics[width=\linewidth]{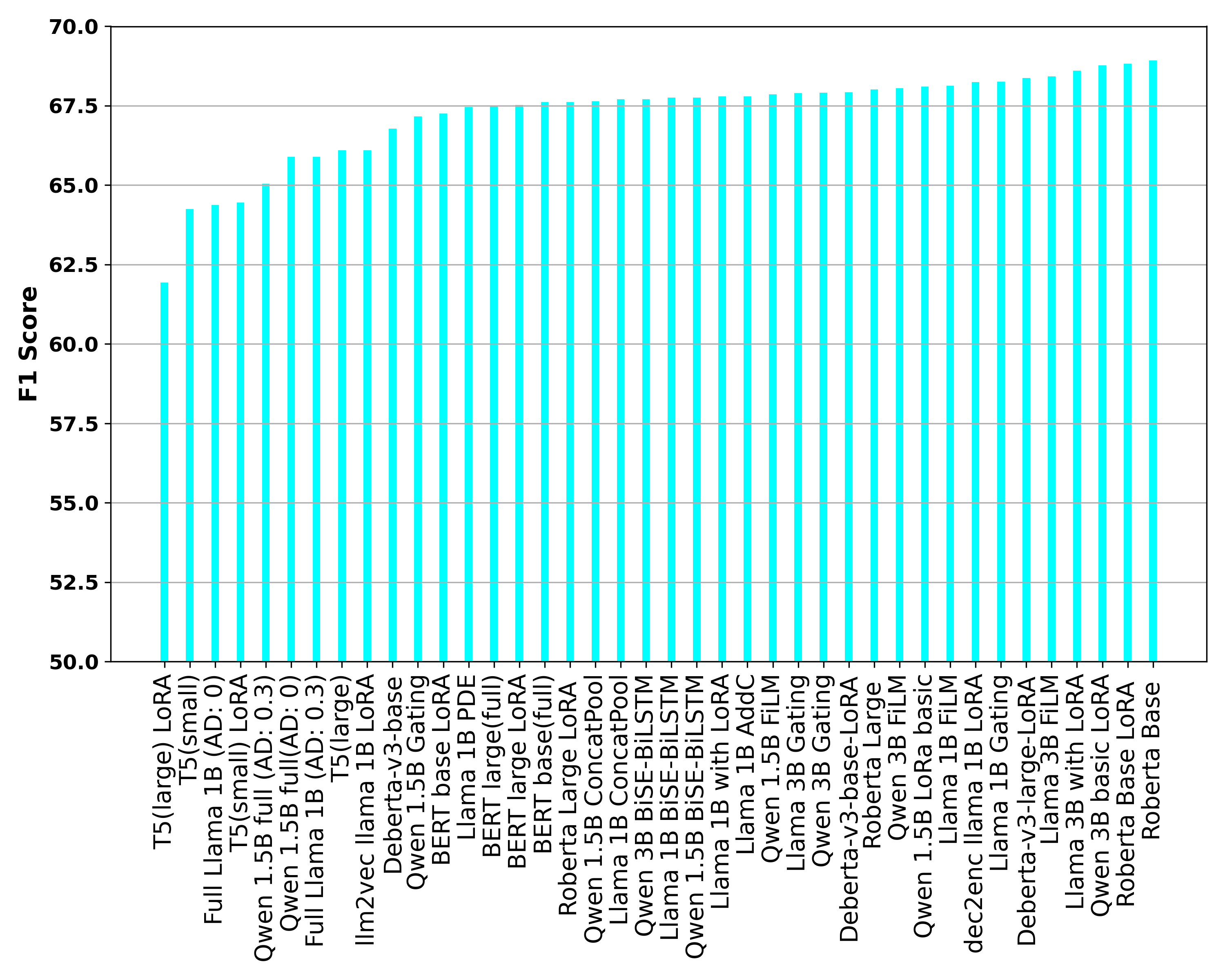}
    \caption{Micro-F1 performance of all fine-tuned models on MAVEN}
    \label{fig:micro-f1}
\end{figure}

Figure~\ref{fig:micro-f1} captures the overall performance by computing global precision and recall across all event types. It is particularly useful when evaluating models on imbalanced datasets, as it favors frequently occurring classes.

\section{Recall for fine-tuning results}
\label{app:recall}
\begin{figure}[ht]
    \centering
    \includegraphics[width=\linewidth]{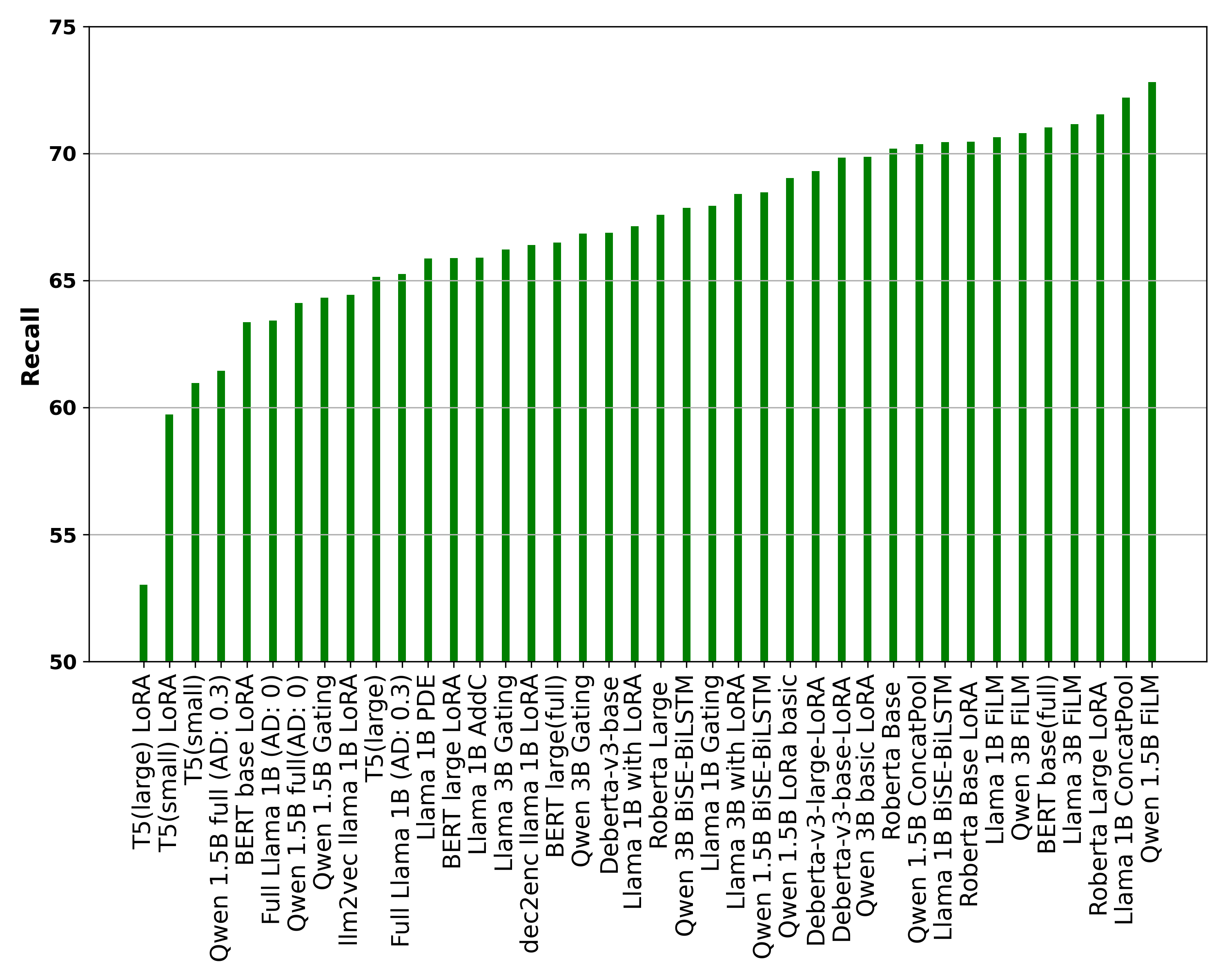}
    \caption{Micro-Recall performance of all fine-tuned models on MAVEN}
    \label{fig:micro-recall}
\end{figure}

\begin{figure}[ht]
    \centering
    \includegraphics[width=\linewidth]{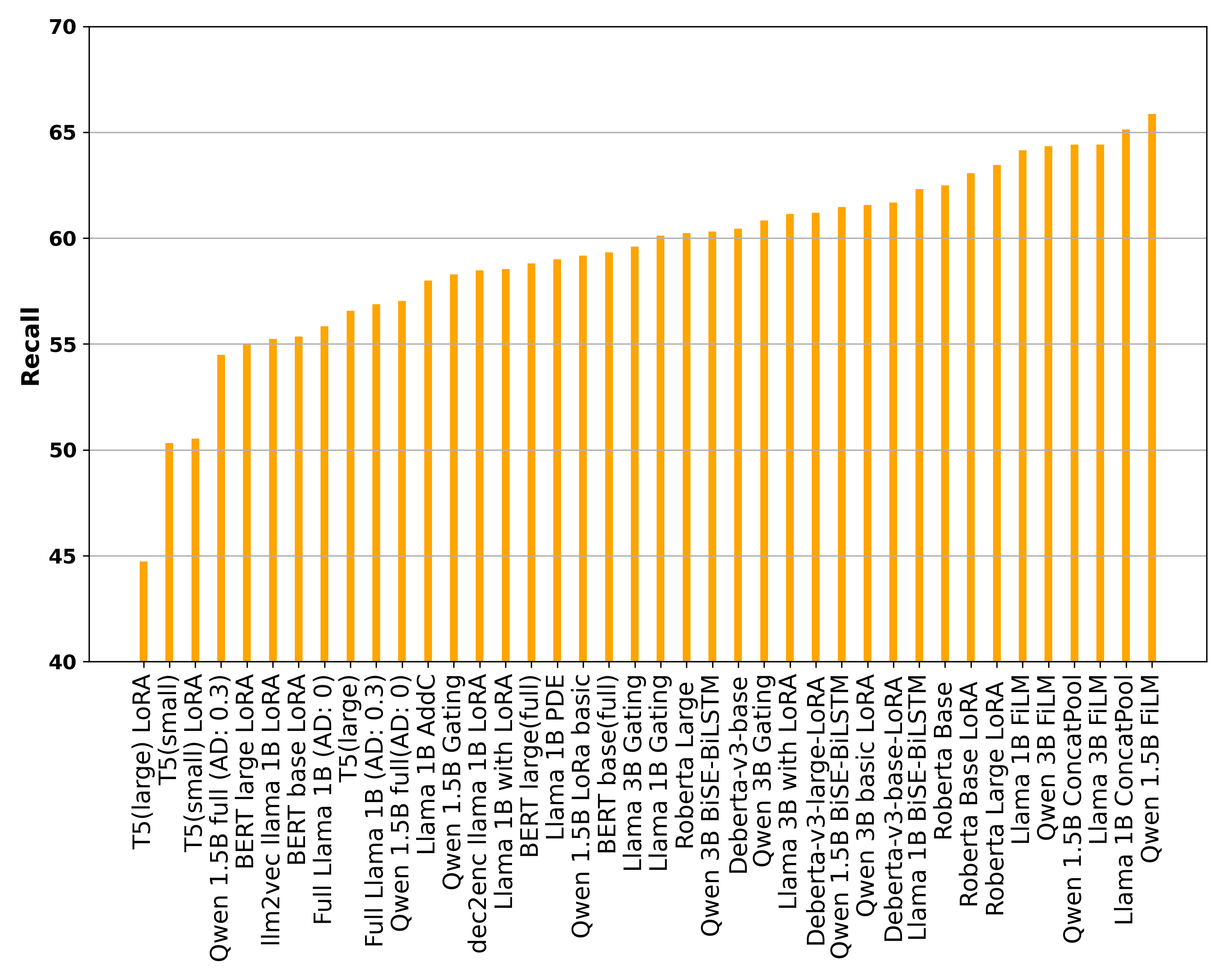}
    \caption{Macro-Recall performance of all fine-tuned models on MAVEN}
    \label{fig:macro-recall}
\end{figure}

Figure~\ref{fig:micro-recall} aggregates true positives across all classes, emphasizing overall sensitivity to detecting events. In contrast, Figure~\ref{fig:macro-recall} treats all classes equally, offering a balanced view of recall performance, especially highlighting how well rare event types are recovered.

\section{Precision for fine-tuning results}
\label{app:precision}
\begin{figure}[ht]
    \centering
    \includegraphics[width=\linewidth]{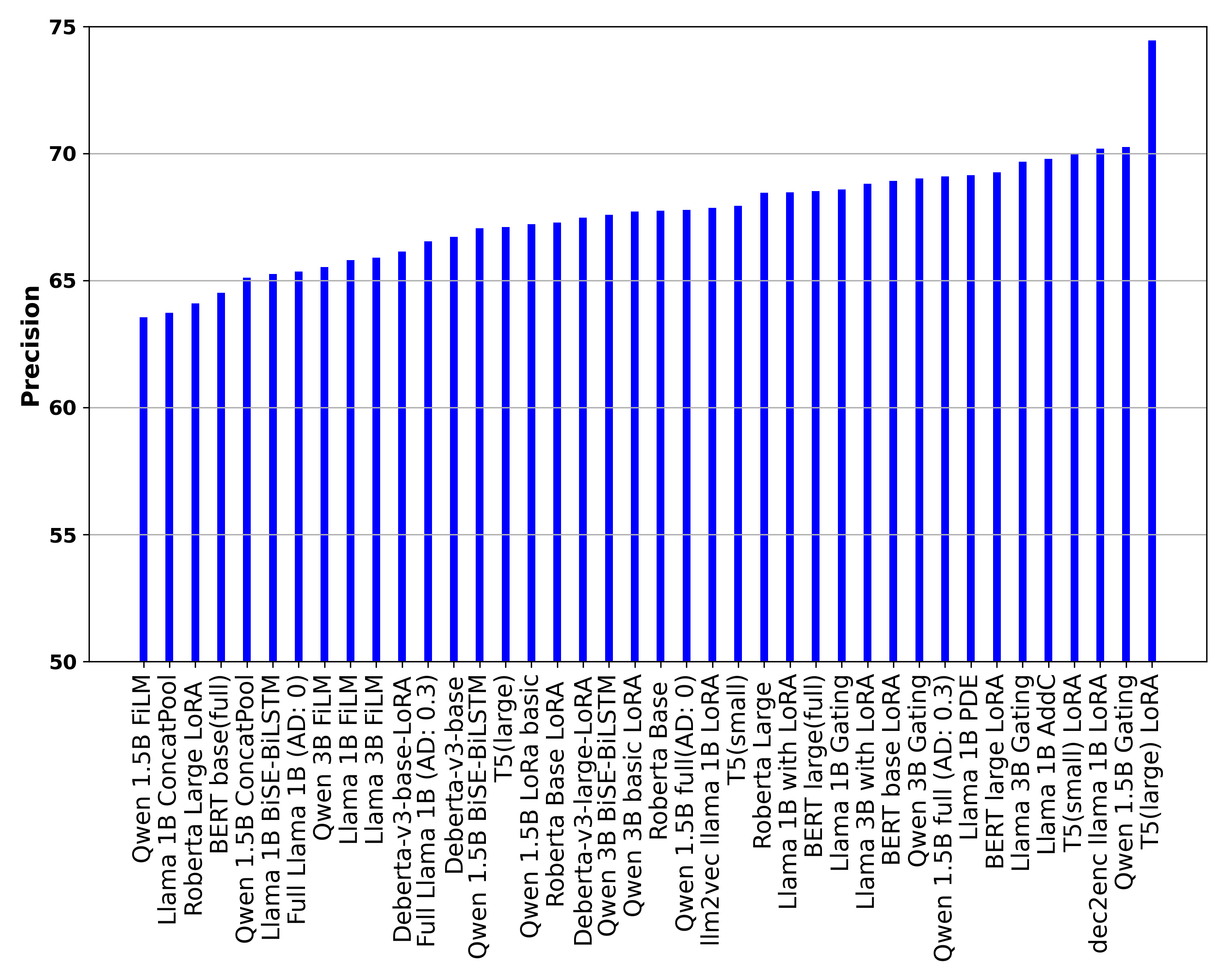}
    \caption{Micro-Precision performance of all fine-tuned models on MAVEN}
    \label{fig:micro-precision}
\end{figure}

\begin{figure}[ht]
    \centering
    \includegraphics[width=\linewidth]{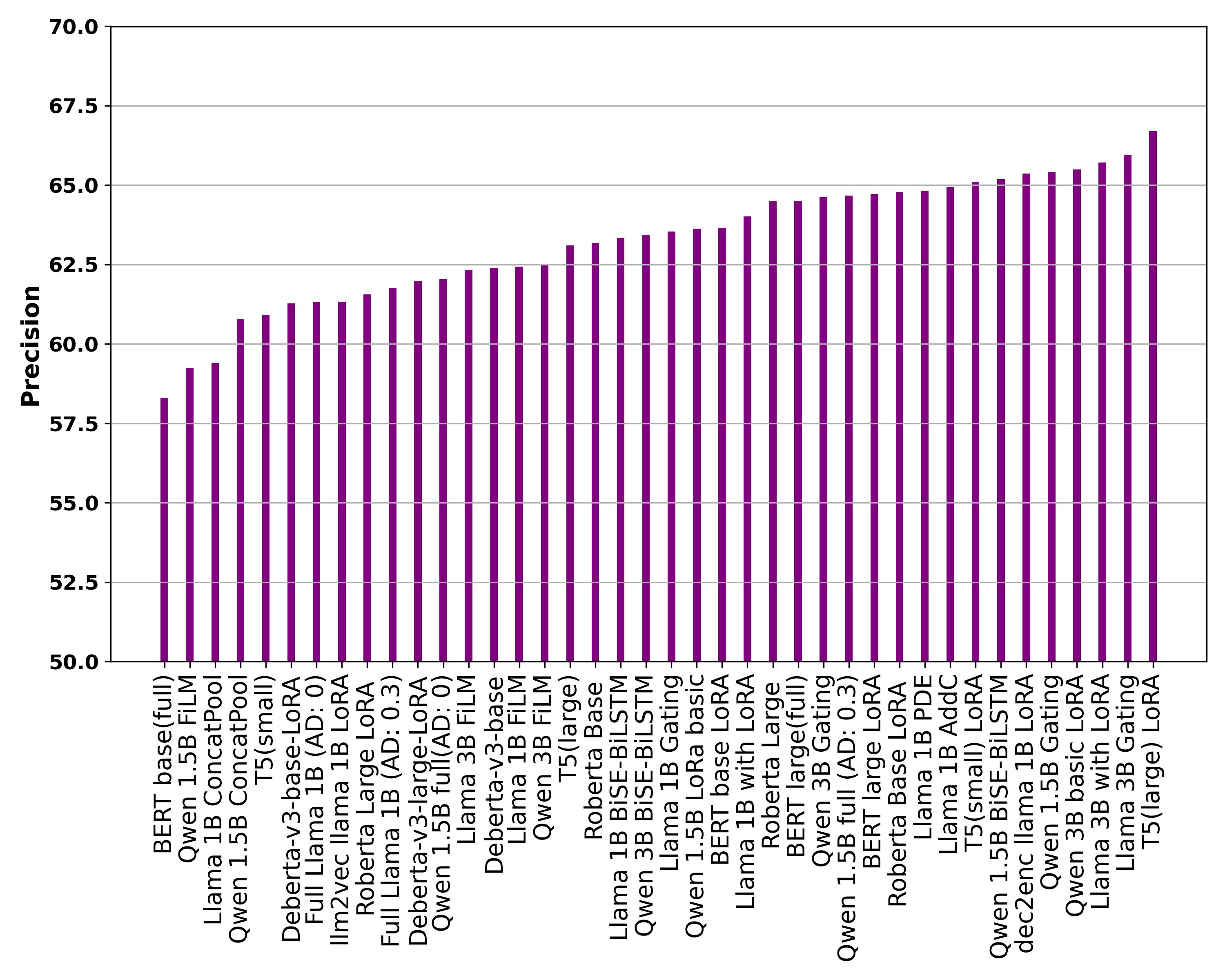}
    \caption{Macro-Precision performance of all fine-tuned models on MAVEN}
    \label{fig:macro-precision}
\end{figure}

Figure~\ref{fig:micro-precision} reflects the model’s ability to avoid false positives across the dataset, while Figure~\ref{fig:macro-precision} averages performance over all event types, revealing how precisely both common and rare events are identified

\section{Testing With Different Splits}
We experimented with the five splits that were used in the TextEE paper (Table \ref{tab:multiple_split}). Also, the average scores are very close to the scores of the split used for all experiments in our paper. These experiments were on the MAVEN dataset.

\begin{table}[t]
\centering
\begin{tabular}{l c}
\hline
\textbf{Model} & \textbf{Micro-F1 (\%)} \\
\hline
Llama full finetuning (our model) & 64.32 \\
Llama FiLM (our model) & 68.04 \\
\hline
\end{tabular}
\caption{Micro-F1 performance comparison of Llama-based models.}
\label{tab:multiple_split}
\end{table}

\section{Metrics According to Sample Frequency}
\label{app:metrics-distribution}
\begin{figure*}[ht]
    \centering
    \includegraphics[width=\textwidth]{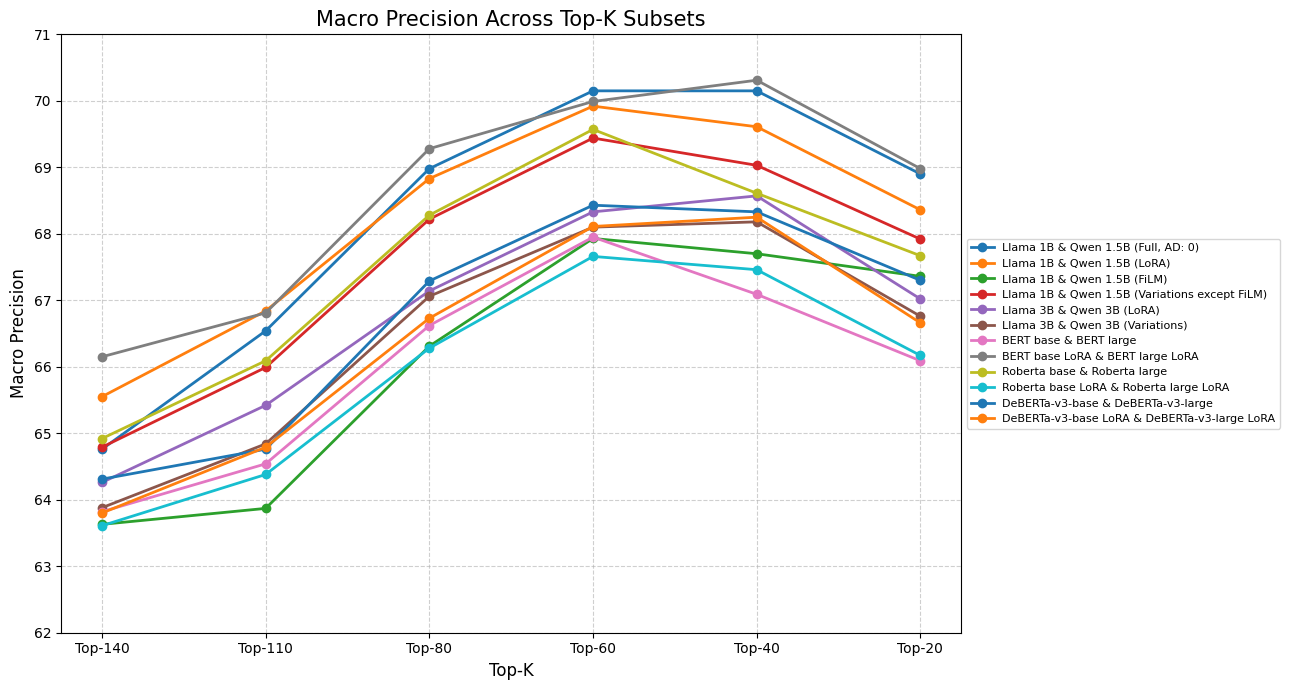}
    \caption{Macro-Precision score distribution across top-k event types}
    \label{fig:macro-precision-dist}
\end{figure*}

\begin{figure*}[ht]
    \centering
    \includegraphics[width=\textwidth]{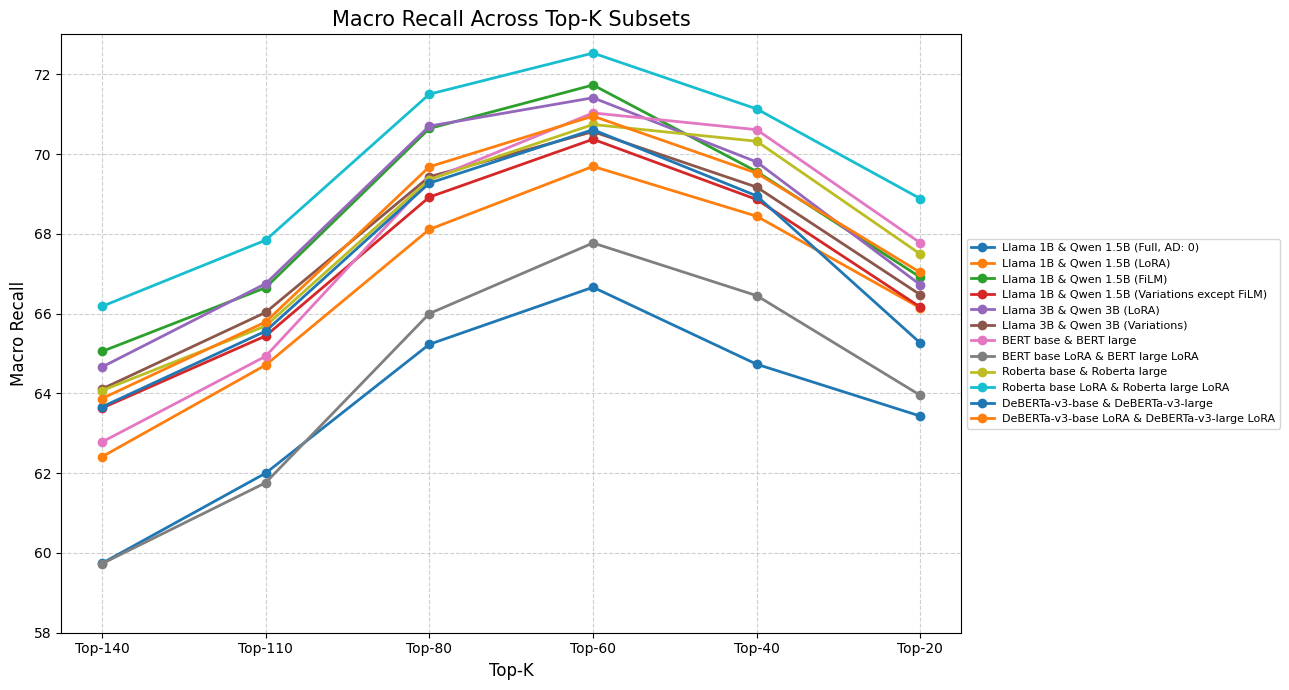}
    \caption{Macro-Recall score distribution across top-k event types}
    \label{fig:macro-recall-dist}
\end{figure*}

\begin{figure*}[ht]
    \centering
    \includegraphics[width=\textwidth]{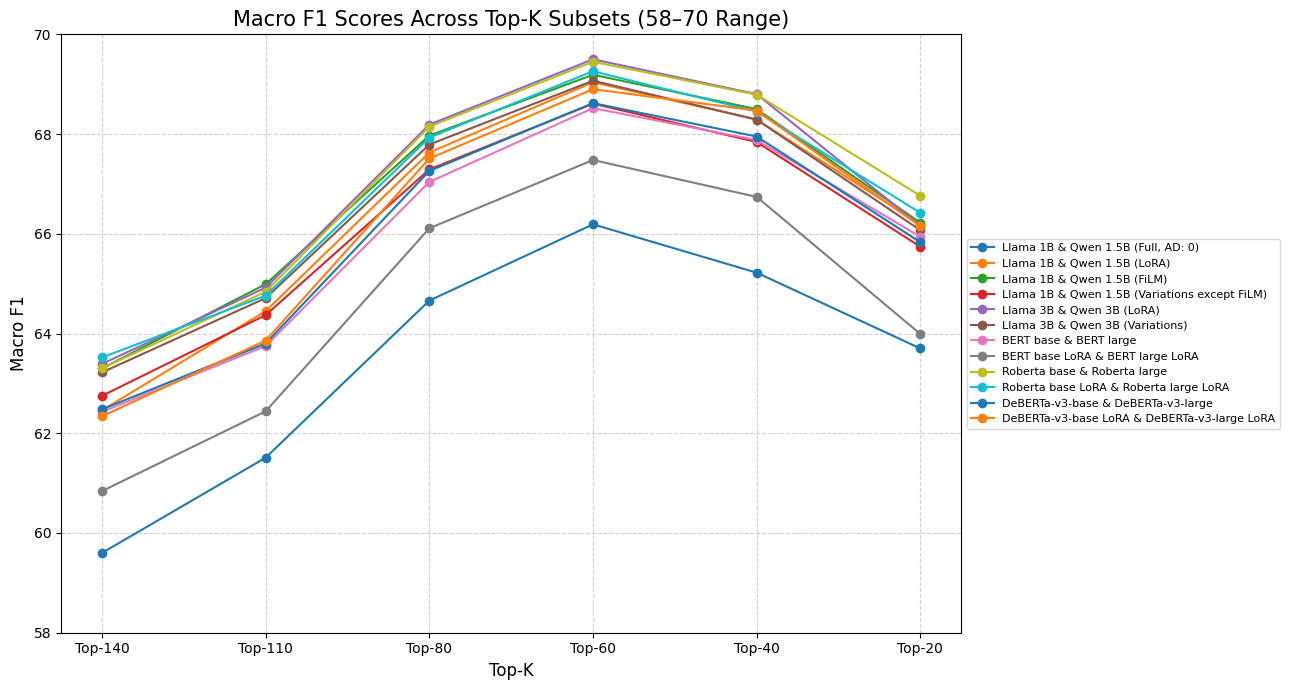}
    \caption{Macro-F1 score distribution across top-k event types}
    \label{fig:macro-f1-dist}
\end{figure*}

Figures \ref{fig:macro-precision-dist}-\ref{fig:macro-f1-dist} show distributions across Top-k event types and provide insight into how model performance scales with class frequency. This highlights whether models generalize well to both high frequency classes and tail classes.

\section{U-shaped Pattern}
Figures ~\ref{fig:macro-precision-dist}, ~\ref{fig:macro-recall-dist}, and ~\ref{fig:macro-f1-dist} show a U-shaped pattern. In addition to event frequency, other factors may influence class performance. For example, among the 40 most frequent event types, some may be closely related to others, which can lead to misclassification.

\section{Impact of LoRA on the Models}
We experimented with multiple ranks of LoRA on the BERT and Llama 1B model. What can be observed from the tables is that LoRA ranks that are higher than rank 8 do not yield additional performance improvements. Experiments on the MAVEN dataset are shown in table X.

\begin{table}[t]
\centering
\footnotesize
\begin{tabular}{lcccc}
\hline
\textbf{Model} & \textbf{LoRA Rank} & \textbf{Micro-F1} & \textbf{Macro-F1} \\
\hline
BERT small & 4  & 66.14 & 55.78 \\
BERT small & 8  & 67.26 & 57.26 \\
BERT small & 16 & 67.18 & 55.01 \\
BERT small & 64 & 56.17 & 66.53 \\
\hline
LLaMA 1B   & 2  & 57.89 & 66.83 \\
LLaMA 1B   & 4  & 58.92 & 67.46 \\
LLaMA 1B   & 8  & 59.17 & 67.58 \\
LLaMA 1B   & 16 & 58.13 & 66.75 \\
\hline
\end{tabular}
\caption{Effect of LoRA rank on Micro-F1 and Macro-F1 for BERT-small and LLaMA-1B models.}
\label{tab:lora_rank_results}
\end{table}

\section{Performance Changes Across Quantiles Induced by LoRA}
After further analyzing the impact of LoRA, we find the following comparison between Llama 1B full finetuning and Llama 1B LoRA finetuning from each quantile. The experiment is done on the MAVEN dataset. The table \ref{tab:quantile_performance_change} shows macro-f1 (\%) improvement from Llama full finetuning to Llama LoRA finetuning. We subtracted the two f1 scores for each quantile.

\begin{table}[t]
\centering
\footnotesize
\resizebox{\columnwidth}{!}{
\begin{tabular}{lcccccc}
\hline
\textbf{Metric} & \textbf{Top-140} & \textbf{Top-110} & \textbf{Top-80} & \textbf{Top-60} & \textbf{Top-40} & \textbf{Top-20} \\
\hline
Performance Change & +3.02 & +3.16 & +3.15 & +2.94 & +2.98 & +2.66 \\
\hline
\end{tabular}
}
\caption{Performance changes across different top-$k$ quantiles}
\label{tab:quantile_performance_change}
\end{table}

As we can see, LoRA increases the F1 scores of least frequent events more than the frequent ones. 

\section{Effect of Upsampling}
\label{app:upsampling}
\begin{table*}[ht]
    \centering
    \small
    \begin{tabular}{lcc}
        \hline
        \textbf{Model} & \textbf{Macro-F1} & \textbf{Micro-F1} \\
        \hline
        BERT (LoRA rank-16) with upsampled dataset    & 48.19 & 65.07 \\
        BERT (LoRA rank-16) no upsample & \textbf{52.99} & \textbf{66.12} \\
        \hline
    \end{tabular}
    \caption{Effect of naive upsampling on BERT rank-16}
    \label{tab:bert-upsampling}
\end{table*}

Naive upsampling does not improve performance as shown in Table~\ref{tab:bert-upsampling}. In fact, both macro and micro F1 scores slightly decrease when upsampling is applied.

\section{Impact of Input Format on Llama}
The table~\ref{tab:lora_input_format} shows that when a preceding space is not used for each word token, the performance of Llama drops significantly. This is because Llama's tokenizer requires a leading space before every word in order to correctly identify word boundaries during tokenization.

\begin{table}[t]
\centering
\begin{tabular}{lccc}
\hline
\textbf{Input Format (Leading Space Tokens)} & \textbf{LoRA} & \textbf{Micro-F1} & \textbf{Macro-F1} \\
\hline
No  & No  & 55.15 & 39.08 \\
No  & Yes & 60.23 & 48.26 \\
Yes & No  & 64.37 & 57.23 \\
Yes & Yes & 67.79 & 59.31 \\
\hline
\end{tabular}
\caption{Impact of input formatting and LoRA on model performance.}
\label{tab:lora_input_format}
\end{table}

\section{Computational Resource and Time}
The GPU configuration is as follows: \\
\textbf{GPU Model:} NVIDIA RTX A6000 \\
\textbf{Architecture:} NVIDIA Ampere \\
\textbf{Memory (VRAM):} 48 GB GDDR6 \\

The table~\ref{tab:training_time_lora} shows the approximate average time for experiments with and without LoRA. LoRA saves both memory and time while finetuning a model.

\begin{table}[t]
\centering
\begin{tabular}{lcc}
\hline
\textbf{Model} & \textbf{LoRA (Rank 8)} & \textbf{Approx. Avg. Time (minutes)} \\
\hline
LLaMA 1B     & Yes & 600 \\
LLaMA 1B     & No  & 900 \\
BERT Small   & Yes & 300 \\
BERT Small   & No  & 420 \\
\hline
\end{tabular}
\caption{Approximate average training time with and without LoRA (rank 8).}
\label{tab:training_time_lora}
\end{table}

\section{Hallucination of Out-of-List Event Types}
\label{app:hallucination}
During evaluation, large language models occasionally hallucinated event types that were absent from the predefined ontology, generating labels not included in the allowed event list. Such out-of-list predictions indicate the model’s tendency to rely on \emph{semantic intuition} rather than strict schema constraints, highlighting a key limitation of open-ended generation. Although these hallucinations were relatively infrequent, they inflated false positives and distorted macro-level metrics. To mitigate this, all outputs were post-validated against the canonical event list, replacing invalid types with a placeholder label such as \texttt{NONE}.